# Lexical Semantics Enhanced Neural Word Embeddings

Dongqiang Yang, Ning Li, Li Zou*, Hongwei Ma*

School of Computer Science and Technology

Shandong Jianzhu University, China

## Abstract

Current breakthroughs in natural language processing have benefited dramatically from- neural language models, through which distributional semantics can leverage neural data representations to facilitate downstream applications. Since neural embeddings use context prediction on word co-occurrences to yield dense vectors, they are inevitably prone to capture more semantic association than semantic similarity. To improve vector space models in deriving semantic similarity, we post-process neural word embeddings through deep metric learning, through which we can inject lexical-semantic relations, including syn/antonymy and hypo/hypernymy, into a distributional space. We introduce hierarchy-fitting, a novel semantic specialization approach to modelling semantic similarity nuances inherently stored in the IS-A hierarchies. Hierarchy-fitting attains state-of-the-art results on the common- and rare-word benchmark datasets for deriving semantic similarity from neural word embeddings. It also incorporates an asymmetric distance function to specialize hypernymy's directionality explicitly, through which it significantly improves vanilla embeddings in multiple evaluation tasks of detecting hypernymy and directionality without negative impacts on semantic similarity judgement. The results demonstrate the efficacy of hierarchy-fitting in specializing neural embeddings with semantic relations in late fusion, potentially expanding its applicability to aggregating heterogeneous data and various knowledge resources for learning multimodal semantic spaces.

## 1. Introduction

Neural language models employ context-predicting patterns rather than the traditional context-counting statistics to yield continuous word embeddings for distributional semantics. Neural word embeddings (NNEs), working either on the character level (Bojanowski et al. 2017) or on the unified (Mikolov et al. 2013a, Mikolov et al. 2013b) vs contextualized (Devlin et al. 2018, Peters et al. 2018) word level, have become a new paradigm for achieving state-of-the-art performances in the benchmark evaluations such as GLUE (Wang et al. 2018) and SuperGLUE (Wang et al. 2019a). Notably, in a broad set of lexical-semantic tasks such as synonym and analogy detection (Baroni et al. 2014), NNEs have significantly improved distributional semantics compared to the traditional co-occurrence counting. For example, after linear vector arithmetic on word2vec (Mikolov et al. 2013a), *queen* was found distributionally close to the composition result of *king – man + woman* in a distributional space.

---

*Co-corresponding author: zouli20|mahongwei@sdjzu.edu.cn.

However, calculating distributional similarity in NNEs usually yields semantic association or relatedness rather than semantic similarity (Hill et al. 2015), inevitably caused by sharing co-occurrence patterns in a context window during self-supervised learning. For example, after calculation of the *cosine* similarity on word embeddings such as the word2vec Skip-gram with Negative Sampling (SGNS) (Mikolov et al. 2013a, Mikolov et al. 2013b), GloVe (Pennington et al. 2014), and fastText (Bojanowski et al. 2017), we find that the most distributionally similar word to *man* is *woman*, and vice versa. In SGNS, *queen* is one of the top 10 similar words to *king*, and vice versa; in GloVe and fastText, *king* is one of the top 10 similar words to *queen*. Although *man* vs *woman* or *king* vs *queen* belongs to antonymy, each pair in the embeddings is scored as highly similar. Semantic relatedness contains various semantic relationships, whereas semantic similarity usually manifests lexical entailment or the IS-A relationship. As hand-crafted knowledge bases (KBs) such as WordNet (Miller 1995, Fellbaum 1998) and BabelNet (Navigli and Ponzetto 2012) mainly consist of IS-A taxonomies, along with synonymy and antonymy, they are often used for computing semantic similarity (Pedersen et al. 2004, Yang and Yin 2021). Distributional semantics needs to fuse semantic relations in KBs to enhance the semantic content in NNEs, which is necessary for improving the generalization of neural language models.

The current study often employs joint-training and post-processing methods to harvest word usage knowledge from distributional semantics and human-curated concept relations from KBs. Most joint-training methods directly impose semantic constraints on their loss functions while jointly optimizing the weighting parameters of neural language models (Yu and Dredze 2014, Nguyen et al. 2017, Alsuhaibani et al. 2018). Another way of joint training is to revise the architecture of neural networks either through training Graph Convolutional Networks with syntactic dependencies and semantic relationships (Vashishth et al. 2019) or by introducing attention mechanisms (Yang and Mitchell 2017, Peters et al. 2019). Joint training can tailor NNEs to specific needs of applications, albeit with an excessive training workload in early fusion. In contrast, the post-processing methods such as retrofitting (Faruqui et al. 2015), counter-fitting (Mrkšić et al. 2016) and LEAR (Vulic and Mrkšić 2018) can avoid such burdensome training processes, semantically specializing NNEs via optimizing a distance metric in late fusion. Semantically enhanced NNEs can facilitate downstream applications, e.g. lexical entailment detection (Nguyen et al. 2017, Vulic and Mrkšić 2018), sentiment analysis (Faruqui et al. 2015, Arora et al. 2020), and dialogue state tracking (Mrkšić et al. 2016, Mrkšić et al. 2017).

Inspired by previous works (Faruqui et al. 2015, Mrkšić et al. 2016, Vulic and Mrkšić 2018) on semantically specializing NNEs in late fusion, we investigate how to post-process NNEs through merging symmetric syn/antonymy and asymmetric hypo/hypernymy. We seek to leverage the IS-A hierarchies' multi-level semantic constraints to augment distributional semantics. By learning distance metrics in a distributional space, we can effectively inject lexical-semantic information into NNEs, pulling similar words closer and pushing dissimilar words further. Consistent results on lexical-semantic tasks show that our novel specialization method can significantly improve distributional semantics in deriving semantic similarity and detecting hypernymy and its directionality.

This paper is organized as follows: Section 2 introduces deep metric learning and examines typical post-processing approaches to injecting semantic relations into neural word embeddings; Section 3 describes hierarchy-fitting, our new late fusion methodology of specializing a distributional space under different semantic constraints; Section 4 outlines our experiments on evaluating hierarchy-fitting, and other popular post-processing approaches in calculating distributional semantics; Section 5 and 6 investigate the efficacy of hierarchy-fitting in refining neural word embeddings through deriving semantic similarity and recognizing hypernymy and its directionality on the benchmark datasets, respectively; Section 7 concludes with several observations and future work.

## 2. Metric learning

The self-supervised training objective of neural language models (NLMs) is to maximize the prediction probability of a token given an input of its context, where cross-entropy is often employed as a cost function for backpropagation to produce NNEs, e.g. word2vec (Mikolov et al. 2013a, Mikolov et al. 2013b) in a simple feedforward network and BERT (Devlin et al. 2018) in a deep transformer network. The joint-training approaches to semantic specialization can directly refine the original training objective with hand-crafted relations (Fried and Duh 2014, Yu and Dredze 2014, Nguyen et al. 2017, Alsuhaibani et al. 2018). To impose semantic constraints on generating neural embeddings, they can also modify the attention mechanisms in recurrent neural networks (Yang and Mitchell 2017) and transformers (Peters et al. 2019). Since the joint-training approaches often produce task-specific NNEs, which are computationally demanding when learning from scratch with massive corpora, we only investigate post-processing approaches that can work on any distributional space.

As with semantic specialization on pre-trained NNEs, instead of cross-entropy loss, ranking loss in deep metric learning (Kaya and BİLge 2019) is often used to learn a Euclidean distance in a latent space under the constraints of semantic relations in KBs. Deep metric learning has broad applications from computer vision (Schroff et al. 2015, Lu et al. 2017) to natural language processing (Mueller and Thyagarajan 2016, Ein Dor et al. 2018, Zhu et al. 2018) to audio speech processing (Narayanaswamy et al. 2019, Wang et al. 2019b). Given two tokens: $x_1$ and $x_2$ in the original vector space of NNEs with a weighting function $f'_\theta: x \in R^n$, metric learning constructs a distance-based loss function $D(f_\theta(x_1), f_\theta(x_2))$ to yield the augmented embeddings with $f_\theta: x \in R^n$. With the help of KBs that specify the relationship between $x_1$ and $x_2$ using a similar or dissimilar tag $y$, metric learning continuously updates $f'_\theta$ to pull similar tokens closer or push dissimilar ones farther, until $D(f_\theta(x_1), f_\theta(x_2))$ finally arrives at minimum for similar tokens and maximum for dissimilar ones.

In deep metric learning, data sampling for computing ranking loss, either in Siamese (Bromley et al. 1993) or Triplet (Hoffer and Ailon 2015) networks, plays a crucial role in specializing neural embeddings. Correspondingly, contrastive or pairwise loss (Chopra et al. 2005) and triplet loss (Schroff et al. 2015) are two popular cost functions in metric learning, followed by many of their variants, such as Quadruple Loss (Ni et al. 2017) and N-Pair Loss (Sohn 2016).

## 2.1 Contrastive loss

Contrastive or pairwise loss (Chopra et al. 2005) was first used for face recognition on the hypothesis that similar faces from the same person should be positioned at a smaller distance in a Euclidean space and different faces from different ones with a larger one. It can be applied in post-processing NNEs as follows:

$$L(x_1, x_2, y) = yD(f_\theta(x_1), f_\theta(x_2)) + (1-y)\max(0, m - D(f_\theta(x_1), f_\theta(x_2)))$$

Here, for the similar tokens: $x_1$ and $x_2$ with the tag $y = 1$, contrastive loss $L$ regards them as a positive sample and seeks to decrease their distance $D(f_\theta(x_1), f_\theta(x_2))$; and for the dissimilar tokens with $y = 0$, it considers them as a negative sample and recommends a distance margin $m$ to regularize $D(f_\theta(x_1), f_\theta(x_2))$. That is to say if $D(f_\theta(x_1), f_\theta(x_2)) > m$, no backpropagation is needed; otherwise, metric learning has to increase their distance. Contrastive loss only works on two token inputs in computing loss every time.

## 2.2 Triplet loss

Triplet loss (Schroff et al. 2015) simultaneously takes three inputs in computing rank loss, which can be defined as follows:

$$L(x_a, x_p, x_n) = \max(0, m + D(f_\theta(x_a), f_\theta(x_p)) - D(f_\theta(x_a), f_\theta(x_n)))$$

For an anchor token $x_a$, $x_p$ and $x_n$ denote its positive and negative samples in a triplet input, respectively. Here, $m$ works as a margin gap to distinguish an easy negative sample from a hard negative one (Kaya and BİLge 2019), and it also serves as a distance boundary between $D(f_\theta(x_a), f_\theta(x_p))$ and $D(f_\theta(x_a), f_\theta(x_n))$ when selecting a triplet in metric learning. If $D(f_\theta(x_a), f_\theta(x_n)) > m + D(f_\theta(x_a), f_\theta(x_p))$, $x_n$ is an easy negative sample as no loss is generated, and it is not necessary to push $x_n$ farther from $x_a$; and if $D(f_\theta(x_a), f_\theta(x_p)) < D(f_\theta(x_a), f_\theta(x_n))$, $x_n$ is a hard negative sample as $x_a$ is distributionally closer to $x_n$ than $x_p$, indicating that backpropagate is needed to update $f_\theta$; and any negative token located between $D(f_\theta(x_a), f_\theta(x_p))$ and $m + D(f_\theta(x_a), f_\theta(x_p))$, is categorized as semi-hard to push away.

In specializing NNEs with prior knowledge, most methods use contrastive loss and triplet loss with different negative-sample selection policies, among which we list some typical ones in the following sections.

## 2.3 Retrofitting

Faruqui et al. (2015) proposed to retrofit word embeddings with semantic lexicons, including PPDB (Ganitkevitch et al. 2013), WordNet (Miller 1995, Fellbaum 1998), and FrameNet (Baker et al. 1998). The positive pair: $x_a$ and $x_p$ should bear a corresponding semantic relationship extracted from lexicons, including lexical paraphrasing in PPDB, synonymy and hypo/pernymy in WordNet, along with words association in FrameNet. These relations were organized into different graphs, in which word embeddings can be altered through belief propagation. The loss function for retrofitting can be articulated as:

$$L(x_a, x_p) = \alpha_a D(f_\theta(x_a), f'_\theta(x_a)) + \sum_{p=1}^{degree(x_a)} \beta_{a,p} D(f_\theta(x_a), f_\theta(x_p))$$

where $\alpha_a$ is often set to 1 to control the specialization strength for $x_a$; and $\beta_{a,p}$, equal to $degree(x_a)^{-1}$, is another regularizing factor for $x_p$ in propagation. Since $L(x_a, x_p)$ is a convex function, minimizing its derivative to $f_\theta(x_a)$ can be denoted as:

$$f_\theta(x_a) = \alpha_a f'_\theta(x_a) + \sum_{p=1}^{degree(a)} \beta_{a,p} f_\theta(x_p) \Big/ \alpha_a + \sum_{p=1}^{degree(a)} \beta_{a,p}$$

Retrofitting works similarly to contrastive loss. Although it extracts multiple positive samples for $x_a$, retrofitting pulls only similar or related tokens closer. Srinivasan et al. (2019) adapted the retrofitting method by introducing a WordNet-based similarity score to better account for the closeness between $x_a$ and its neighbours that are located within a 2-link distance in an IS-A hierarchy, and achieved competitive results in intrinsic and extrinsic evaluations.

*2.4 Counter-fitting.*

Inspired by retrofitting, Mrkšić et al. (2016) incorporated synonymy and antonymy in semantically enhancing word embeddings. They linearly assembled the loss functions from different semantic constraints while preserving distributional semantics, which are:

1. Synonymy: $L(x_a, x_p) = \max\left(0, m_{syn} - D\left(f_\theta(x_a), f_\theta(x_p)\right)\right)$

2. Antonymy: $L(x_a, x_n) = \max\left(0, D(f_\theta(x_a), f_\theta(x_n)) - m_{ant}\right)$

3. Distributional semantics:

$$L(x_a, x_j) = \sum_{j \in N(x_a)} max\left(0, D\left(f_\theta(x_a), f_\theta(x_j)\right) - D\left(f'_\theta(x_a), f'_\theta(x_j)\right)\right)$$

Mrkšić et al. (2016) set up different loss functions for synonymy and antonymy and sequentially specialized NNEs. $m_{syn}$ and $m_{ant}$ are margins for synonymy and antonymy, respectively. Besides semantic specialization on NNEs, they also preserved distributional semantics using $L(x_a, x_j)$, where $x_j$ is one of the top distributionally similar words to $x_a$ in $N(x_a)$. In place of the whole vocabulary with $N(x_a)$, $x_j$ also acts as a pseudo-negative word in $L(x_a, x_j)$ for efficient backpropagation. Note that the Euclidean distance $D$ is often converted with the *cosine* similarity. Counter-fitting bears a close resemblance to contrastive loss as synonymy and antonymy constraints specialize NNEs conversely.

*2.5 ATTRACT-REPEL.*

Mrkšić et al. (2017) further improved counter-fitting with semantic constraints from mono- and cross-lingual resources. They used triplet loss rather than contrastive loss to refine a distributional space, i.e. attracting synonyms and repelling antonyms, therefore termed ATTRACT-REPEL. The loss functions in ATTRACT-REPEL are listed as follows:

1. Synonymy (ATTRACT):

$$L(x_a, x_p, x_{ns}) = \sum_{x_{ns}} max(0, m_{syn} + D(f_\theta(x_a), f_\theta(x_p)) - D(f_\theta(x_a), f_\theta(x_{ns})))$$

2. Antonymy (REPEL):

$$L(x_a, x_n, x_{ps}) = \sum_{x_{ps}} max\left(0, m_{ant} - D(f_\theta(x_a), f_\theta(x_n)) + D(f_\theta(x_a), f_\theta(x_{ps}))\right)$$

3. Distributional semantics:

$$L(x_a, x_p, x_n) = m_{reg}(D(f'_\theta(x_a), f_\theta(x_a)) + D(f'_\theta(x_p), f_\theta(x_p)) + D(f'_\theta(x_n), f_\theta(x_n)))$$

While injecting semantic constraints into a vector space, ATTRACT-REPEL retrieved online samples from each mini-batch to calculate the Euclidean distance $D$. Specifically for $x_a$ in the triplet $(x_a, x_p, x_{ns})$, to pull synonyms closer, the negative sample, $x_{ns}$, is one of the remaining tokens in a mini-batch that holds the shortest distance to $x_a$. Likewise, to push antonyms farther, the positive sample, $x_{ps}$, is one of the remaining tokens in a mini-batch that holds the longest distance to $x_a$. Since synonymy and antonymy are both semantically symmetrical, ATTRACT-REPEL also takes $x_p$ and $x_n$ as anchor nodes to $x_a$ to recalculate corresponding triplet losses, respectively. Except for the positive and negative samples ($x_p$ and $x_n$) to $x_a$, there is no other online selection of samples in preserving distributional semantics.

After ATTRACT-REPEL, the specialized NNEs achieved better outcomes than counter-fitting in similarity judgement and dialogue state tracking tasks. Moreover, thanks to cross-lingual links in the multilingual KBs such as PPDB (Ganitkevitch et al. 2013) and BabelNet (Navigli and Ponzetto 2012), ATTRACT-REPEL can enhance word embeddings in other languages that lack lexical resources.

*2.6 LEAR.*

Apart from syn/antonymy in ATTRACT-REPEL, LEAR (Vulic and Mrkšić 2018) also incorporated lexical entailment or hypernymy to refine NNEs. Given that lexical entailment in WordNet is organized into a hierarchy, semantic constraints in LEAR consist of direct or immediate hypernymy and indirect one that holds more than a two-link distance. LEAR established the same distance margin for hypernymy and synonymy to pull any hyponym–hypernym pair closer. Moreover, LEAR employed asymmetric distance metrics to learn hypernymy's directionality, assuming that a concept's vector magnitude should be less than its hypernym's after semantic specialization. As a variant of ATTRACT-REPEL, LEAR firstly defined the symmetric ATTRACT cost on hypernymy constraints as follows:

$$L(x_a, x_p, x_{ns}) = \sum_{x_{ns}} max\left(0, m_{syn} + D(f_\theta(x_a), f_\theta(x_p)) - D(f_\theta(x_a), f_\theta(x_{ns}))\right)$$

It then quantified an asymmetric distance metric to encode hypernymy's directionality. LEAR converted the dot product of $D$ in ATTRACT-REPEL with the *cosine* similarity. Since LEAR singled out the directionality of lexical entailment as a separate learning objective, it achieved state-of-the-art results in recognizing hypernymy.

*2.7 LexSub*

Instead of fusing various semantic constraints to specialize a unified distributional space, e.g. in ATTRACT-REPEL (Mrkšić et al. 2017) and LEAR (Vulic and Mrkšić 2018),

LexSub (Arora et al. 2020) learned a separate projection matrix to construct a subspace for each semantic constraint. Semantic constraints in LexSub consist of symmetric syn/antonymy, asymmetric IS-A relations or hypo/hypernymy, and asymmetric PART-OF relations or mero/holonymy. Arora et al. (2020) claimed that the main advantage of LexSub over other post-processing methods was that it might avoid the interaction of different semantic relationships during specialization and be particularly helpful for some domain-specific applications. To learn a projected subspace for each semantic constraint, LexSub employed contrastive loss rather than the triplet loss in ATTRACT-REPEL and LEAR, which can be defined as follows:

1. Symmetric ATTRACT of synonymy:

$$L(x_a, x_p, x_{ns}) = D\left(f_\theta^{syn}(x_a), f_\theta^{syn}(x_p)\right) + \mu * \sum_{ns \in batch} max\left(0, m_{syn} - D\left(f_\theta^{syn}(x_a), f_\theta^{syn}(x_{ns})\right)\right)$$

2. Symmetric REPEL of antonymy:

$$L(x_a, x_n, x_{ps}) = D\left(f_\theta^{ant}(x_a), f_\theta^{ant}(x_n)\right) + \mu * \sum_{ps \in batch} max\left(0, D\left(f_\theta^{ant}(x_a), f_\theta^{ant}(x_{ps})\right) - m_{ant}\right)$$

3. Asymmetric ATTRACT of hypernymy or meronymy:

$$L(x_a, x_p) = D(f_\theta^{hyp}(x_a), x_p) + max\left(0, \gamma - D(f_\theta^{hyp}(x_p), x_a)\right) + \mu * \sum_{sna \in batch} max\left(0, m_{hyp} - D(f_\theta^{hyp}(x_a), x_{sna})\right)$$

Here, $\mu$ and $\gamma$ are two tunable hyperparameters, and each loss function works in a separate subspace permutated by a projection matrix $f$. For example, for the symmetric ATTRACT of synonymy, $f_\theta^{syn}$ transpose a data point $x$ in the original neural embeddings of $f_\theta': x \in R^n$ into $f_\theta^{syn}: x \in R^d$, using a learned $d$ by $n$ matrix $W^{syn}$, i.e. $f_\theta^{syn}(x) = W^{syn} f_\theta'(x)$.

A key difference of LexSub from the triplet-loss based methods, including ATTRACT-REPEL and LEAR, is negative sample selection. LexSub chose a group of negative samples $sna$ in metric learning under the condition that $D(f_\theta^{syn}(x_a), f_\theta^{syn}(x_{sna})) < m_{syn}$ or $D(f_\theta^{ant}(x_a), f_\theta^{ant}(x_{sna})) > m_{ant}$.

As for the asymmetric relations, LexSub also applied affine transformation on $x_a$ to learn an asymmetric distance metric between hyponym and hypernym in determining their directionality. LexSub significantly improved the performance of hypernym detection and directionality in comparison with other post-processing methods such as retrofitting, counter-fitting, and LEAR. Note that LexSub also retrofitted VSMs with asymmetric mero/holonymy, which indicates that the refined vector space may be suitable for computing semantic relatedness.

## 2.8 Summary

Overall, apart from enhancing NNEs with different semantic constraints, the main differences among the late fusion approaches lie in choosing loss functions and data sampling, as shown in Table 1.

Post-processing methods mainly focus on metric learning to semantically specialize a distributional space, but coupling positive and negative samples in metric learning is also widely employed in self-supervised training. For example, to maximize the likelihood of word prediction in word2vec (Mikolov et al. 2013a), except for the use of hierarchical softmax, negative sampling (Mikolov et al. 2013b) simplified the optimization process through constructing a contrastive loss, preventing it from learning identical embeddings (Goldberg and Levy 2014), which can be formulated as:

$$L(w,c,n) = log(\sigma(f_\theta(w) \odot f'_\theta(c))) + \sum_n log\ (\sigma(-f_\theta(w) \odot f'_\theta(n)))$$

Table 1: Summarization of different post-processing methods for specializing NNEs. Syn, Ant, Hyper, and Mero denote synonymy, antonymy, hypernymy, and meronymy, respectively.

|  | Loss | Negative Sampling | Semantic constraints | Hyper-directionality |
|---|---|---|---|---|
| Retrofitting | Distance | Null | Syn | Null |
| Counter-fitting | Contrastive | Null | Syn+Ant | Null |
| ATTRACT-REPEL | Triplet | Online selection | Syn+Ant | Null |
| LEAR | Triplet | Online selection | Syn+Ant+Hyper | Vector norms |
| LexSub | Contrastive | Online selection | Syn+Ant+Hyper+Mero | Affine transformation |

Here, *w* and *c* denote a word and its neighbour in context, respectively; n is one of the negative samples randomly retrieved in a vocabulary for loss calculation. $f_\theta$ and $f'_\theta$ stand for the weight matrices learned for the hidden and output layers in SGNS, respectively. $\sigma$ denotes the *sigmoid* function to derive the prediction probability of *c* occurrence given the target word *w* in skip-gram, and the dot product of vectorized words indicates their distributional relatedness. Hence, the objective of the loss function in SGNS is two-fold: maximizing the probabilities between *w* and its contextual words (positive) inside a sliding window and simultaneously minimizing the probabilities between *w* and a negative sample randomly selected from the outside of the window. It can prevent Skip-gram or CBOW from updating all the weights in the output layer and significantly lower the loss calculation complexity incurred by *softmax*. Therefore, the objective of the negative sampling in the above loss function is in line with it in the contrastive loss of metric learning, which aims to attract related words while repelling unrelated ones in a Euclidean space. Note that in learning hierarchical embeddings via joint training in early fusion, Nguyen et al. (2017) employed a triplet distance metric and selected distinctive contexts for hypo/hypernym to minimize their distributional difference, which works equivalently to injecting lexical entailment into a distributional space.

Moreover, the log probability in $L(w,c,n)$ that inherently captures distributional similarity plays a similar role as a distance metric in late fusion. In measuring taxonomic similarity, Resnik (1995) proposed to compute information content instead of simple edge-counting in an IS-A hierarchy, given the semantic variation of a single link in the hierarchy. The information content of a conceptual node is equal to the negative log probability derived from summing up the occurrences of its hyponymy children (both direct and indirect) in a corpus. Therefore, as a conceptual node ascends in a hierarchy, its probability increases and its information content decreases correspondingly, indicating that it provides less helpful information for computing semantic similarity because it becomes more abstract or general. So given a pair of words: $w_1$ and $w_2$, their taxonomic similarity can be defined as $Sim(w_1, w_2) = -log(ncn(w_1, w_2))$, where $ncn$ is the nearest common node for $w_1$ and $w_2$ in an IS-A hierarchy, and information content can replace standard edge-counting methods with corpus statistics in calculating taxonomic similarity. Likewise, the log probability in $L(w, c, n)$ is practically the same as information content, which can derive semantic similarity through calculating distributional similarity.

## 3. Hierarchy-fitting

Under the hypothesis of similar words sharing similar contexts in vector semantics (Harris 1954, Firth 1957), words that frequently co-occur in a context window will result in similar vectors. For example, the training objective of SGNS is to maximize the similarity of a word with its contextual words and minimize it with negative samples, which will inevitably yield similar embeddings for highly associated words in context (Goldberg and Levy 2014). Therefore, distributional semantics derived from neural embeddings may mix semantic relatedness with semantic similarity (Hill et al. 2015, Lê and Fokkens 2015), in which synonyms are hardly distinguishable from antonyms.

We propose post-processing NNEs to enhance distributional semantics in deriving semantic similarity rather than relatedness. We collect relationships primarily used to compute taxonomic similarity, including syn/antonymy and hypo/hypernymy, to refine NNEs. Instead of learning a separate subspace for each type of relationship in LexSub (Arora et al. 2020), we follow the same procedure as ATTRACT-REPEL (Mrkšić et al. 2017) and LEAR (Vulic and Mrkšić 2018) in combining different relationships to construct a unified VSM space. The main drawbacks of LexSub exist in its extra cost of learning another layer of weights for each semantic constraint to project an original distributional space into a dedicated one. We construct a separate loss function for each semantic constraint and correspondingly run AdaGrad with a single mini-batch for backpropagation. It can effectively lower the interaction of specialization with different semantic constraints.

### 3.1 Injecting Synonymy and antonymy

We impose specialization on a distributional space with two key semantic constraints in computing semantic similarity: synonymy and antonymy. Their cost functions can be outlined as follows:

1. ATTRACT with synonymy

$$L(x_a, x_{syn}, x_{ns}) = \sum_{x_{ns}} max\left(0, m_{syn} + D\left(f_\theta(x_a), f_\theta(x_{syn})\right) - D\left(f_\theta(x_a), f_\theta(x_{ns})\right)\right)$$

2. REPEL with antonymy

$$L(x_a, x_{ant}, x_{ps}) = \sum_{x_{ps}} max\left(0, m_{ant} + D\left(f_\theta(x_a), f_\theta(x_{ps})\right) - D\left(f_\theta(x_a), f_\theta(x_{ant})\right)\right)$$

For the synonymy specialization in the triplet loss $L(x_a, x_{syn}, x_{ns})$, the margin $m_{syn}$ imposes a constraint on the range of distributional distance among the anchor $x_a$, its synonym $x_{syn}$, and its negative sample $x_{ns}$, namely, pulling $x_a$ and $x_{syn}$ closer and pushing $x_a$ and $x_{ns}$ farther. As for $L(x_a, x_{ant}, x_{ps})$ for antonymy, $m_{ant}$ serves in constraining $x_a$, its positive sample $x_{ps}$, and its antonym $x_{ant}$. Given the symmetric feature of synonymy and antonymy, we also apply the same hinge loss function on the triplets of $(x_{syn}, x_a, x_{ns})$ and $(x_{ant}, x_a, x_{ps})$, respectively, in each mini-batch. To better regularize the Euclidean distance between $x_a$ and its synonym in computing loss gradient, we mainly retrieve informative negative samples from a remaining mini-batch, which are distributionally closest tokens to $x_a$ or $x_{syn}$. Likewise, the positive samples for antonymy consist of tokens that are distributionally farthest to $x_a$ or $x_{ant}$ in a remaining mini-batch. We convert the *cosine* similarity into the vector distance *D*.

*3.2 Retrofitting hypernymy*

Apart from injecting syn/antonymy into distributional semantics, we also leverage lexical entailment or the IS-A hierarchical relationship in specialization, which works as:

$$L(x_a, x_{hyp}, x_{ns}) = \sum_{x_{ns}} max\left(0, m_{hyp} + D\left(f_\theta(x_a), f_\theta(x_{hyp})\right) - D\left(f_\theta(x_a), f_\theta(x_{ns})\right)\right)$$

Since hyponym−hypernym pairs are less similar than synonymous ones in weighting lexical-semantic relations (Hirst and St-Onge 1997), we set up a different margin of $m_{hyp}$ from $m_{syn}$ to regularize the Euclidean distances of $x_a$ to its direct hypernym $x_{hyp}$ and negative sample $x_{ns}$. Note that we only employ the direct hypernym of $x_a$ in computing the cost function, which is different from LEAR that retrieves both direct and indirect hypernyms of $x_a$. We conjecture that imposing a unanimous distance margin on the hyponym–hypernym pairs that are located on different levels of IS-A hierarchy may impose inaccurate semantic constraints on specialization, causing miscalculating the loss gradient.

Assuming that a concept's vector norm or magnitude should be shorter than its hypernym's, LEAR also complemented its specialization objective with an asymmetric distance function to detect hypernymy and directionality in distributional semantics. We suggest that encoding the directionality of lexical entailment through vector norms would further interfere with the triplet loss function. Since our primary goal of specializing neural embeddings in late fusion is to improve their capability of deriving semantic similarity, we neglect the asymmetric distance function at the current phrase. We only use it for detecting hypernymy and directionality in a distributional space.

*3.3 Adjusting hierarchy relationship*

Apart from using the triplet loss functions to inject semantic constraints, we define a quadruplet loss (Chen et al. 2017) to regularize further semantic variation in the IS-A hierarchy, hence termed hierarchy-fitting, which is defined as:

$$\begin{aligned}L(x_a, &x_{syn}, x_{hyp}, x_{ns}) \\= &\ max(0, m_{hie-syn} + D(f_\theta(x_a), f_\theta(x_{syn})) - D(f_\theta(x_a), f_\theta(x_{hyp}))) \\&+ \sum_{x_{ns}} max(0, m_{hie-hyp} + D(f_\theta(x_a), f_\theta(x_{syn})) - D(f_\theta(x_{hyp}), f_\theta(x_{ns}))) \\&+ max(0, m_{hie-syn} + D(f_\theta(x_{syn}), f_\theta(x_a)) - D(f_\theta(x_{syn}), f_\theta(x_{hyp}))) \\&+ \sum_{x_{ns}} max(0, m_{hie-hyp} + D(f_\theta(x_{syn}), f_\theta(x_a)) - D(f_\theta(x_{hyp}), f_\theta(x_{ns})))\end{aligned}$$

Here $m_{hie\text{-}syn}$ and $m_{hie\text{-}hyp}$ are two margins assigned for synonymy and hypernymy, respectively, to fine-tune their semantic nuances. Unlike $m_{syn}$ or $m_{hyp}$ in the triplet loss, $m_{hie\text{-}syn}$ attempts to differentiate synonymy from hypernymy using a tiny distance margin to show their nuances in measuring semantic similarity. Given an anchor word $x_a$, its synonym $x_{syn}$ and hypernym $x_{hyp}$ bears a high semantic resemblance to $x_a$ but having some degree of semantic variations, namely, $x_a$ is semantically closer to $x_{syn}$ than to $x_{hyp}$. In the quadruplet loss, $m_{hie\text{-}syn}$ imposes a constraint on the range of distributional distances among $x_a$, $x_{syn}$, and $x_{hyp}$, namely, attracting $x_a$ and $x_{syn}$ closer and repelling $x_a$ and $x_{hyp}$ farther. After that, $m_{hie\text{-}hyp}$ further regularizes distributional distances among the quadruplet of $x_a$, $x_{syn}$, $x_{hyp}$, and $x_{ns}$, where $x_{ns}$ is a negative sample for optimizing loss computation. In line with the synonymy specialization, we retrieve $x_{ns}$ that is distributionally closest to $x_a$ or $x_{syn}$ in the remaining minibatch for backpropagation. Correspondingly, we enforce an identical procedure to calculate another quadruplet loss for $x_{syn}$.

Unlike LEAR and LexSub, which treat synonymy and hypernymy identically on the semantic specialization of embeddings, we propose hierarchy-fitting to differentiate them, as illustrated in Figure 1. Hierarchy-fitting can fine-tune similarity variation between synonyms and hyponym–hypernyms via explicitly weighting hierarchical information in a quadruplet loss. Intuitively, the two margins: $m_{hie\text{-}syn}$ and $m_{hie\text{-}hyp}$ are supposed to regularize distributional distances of $x_a$ to $x_{syn}$, $x_{hyp}$, and $x_{ns}$, namely

$$D(f_\theta(x_a), f_\theta(x_{syn})) \leq D(f_\theta(x_a), f_\theta(x_{hyp})) \leq D(f_\theta(x_a), f_\theta(x_{ant}))$$

It can preserve hierarchical information between concepts. Since not every word can find its antonyms, we replace them with negative samples retrieved in a remaining minibatch during training.

Sample selection is critical in speeding up convergence and reducing the odd of local minima (Kaya and BİLge 2019). As depicted in Figure 1, as with the Euclidean distance to $x_a$, negative samples in hierarchy-fitting include:

1. Hard ones with their distances less than $D(f_\theta(x_a), f_\theta(x_{syn}))$;

2. Semi-hard ones with their distances between $D(f_\theta(x_a), f_\theta(x_{syn}))$ and $m_{hie\text{-}syn}+D(f_\theta(x_a), f_\theta(x_{syn}))$;

3. Easy ones with their distances larger than surpassing $m_{hie\text{-}syn}+D(f_\theta(x_a), f_\theta(x_{syn}))$.

Hard negative samples may cause a high variance in computing the loss function gradient, whereas easy ones may contribute little to refining models. Hard, semi-hard, and easy sampling have applications in deep metric learning (Kaya and BİLge 2019), with hard and semi-hard ones more informative for training. As for sampling strategies, we found that retrieving two negative samples in LEAR: one is randomly selected, and the other is distributionally closest to the anchor works effectively in our experiments.

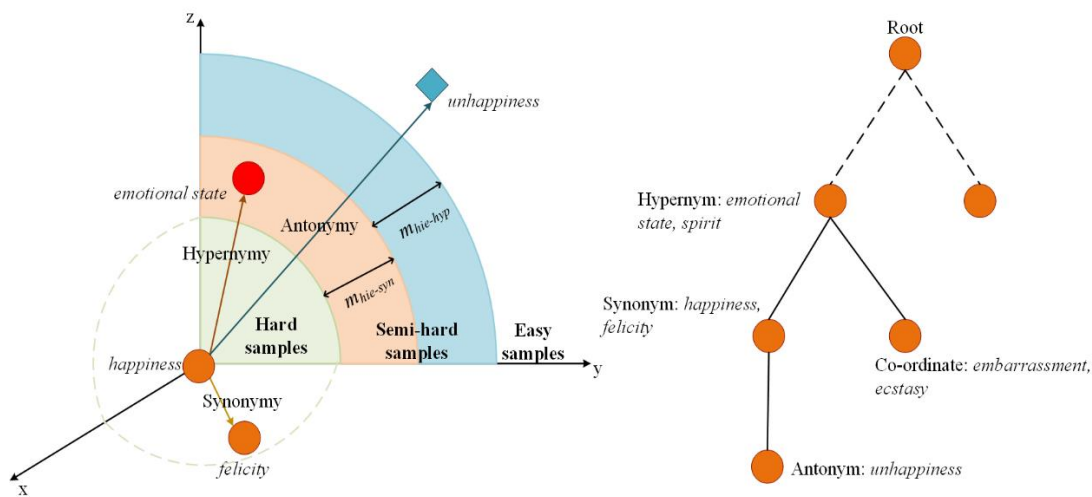

Figure 1. A hypothetical illustration of semantic specialization with a quadruplet loss. Given an IS-A hierarchy of WordNet starting from the root node on the right, each node depicts a synset that usually contains a group of synonyms. A synset of (*happiness*, *felicity*) has an antonym node of (*unhappiness*), a direct hypernym node of (*emotional state, spirit*), and a coordinate node of (*embarrassment, ecstasy*). The injection process of these semantic relations into neural embeddings is depicted on the left. The quadruplet loss first fine-tunes the distances of *happiness* to *felicity* and *emotional state* with the margin $m_{hie\text{-}syn}$, and then to *emotional state* and *unhappiness* with the margin $m_{hie\text{-}hyp}$.

### 3.4 Preserving vector space

Besides injecting semantic relationships into a distributional model, we also preserve distributional semantics learned from co-occurrence counts in context, which can be defined as:

$$L_{dist} = \gamma_{reg} \sum_{x \in batch} D\big(f_\theta(x), f'_\theta(x)\big)$$

We set up a regularizing factor, $\gamma_{reg}$, to fine-tune embeddings in fusing semantic relations hand-crafted in KBs and distributional semantics harvested from text corpora.

### 3.5 Loss in total

Finally, the overall loss function of semantic specialization on a distributional space sums up the above five costs, i.e. $L(x_a, x_{syn}, x_{ns})$, $L(x_a, x_{ant}, x_{ps})$, $L(x_a, x_{hyp}, x_{ns})$, $L(x_a, x_{syn}, x_{hyp}, x_{ns})$, and $L_{dist}$, to enrich distributional semantics with lexical semantics for semantic similarity computation.

## 4. Experiments

### 4.1 Word embeddings

To thoroughly investigate hierarchy-fitting along with other post-processing methods, we incorporated a group of popular word embeddings in the experiments, including Skip-gram with Negative Sampling (SGNS) (Mikolov et al. 2013a), GloVe (Pennington et al. 2014), and fastText (Bojanowski et al. 2017). These embeddings usually assume a bag-of-word (SGNS and GloVe) or bag-of-character (fastText) context to collect co-occurrence statistics. Note that GloVe explicitly factorizes a word by context matrix to reduce its dimensionality in yielding condensed embeddings, whereas SGNS implicitly factorises a co-occurrence counts matrix (Levy and Goldberg 2014). They may be envisioned as prediction-based or neural embeddings, as Levy et al. (2015) suggested.

We collected the publicly available English embeddings for SGNS, GloVe, and fastText, with the same dimensionality of 300 to avoid the bias caused by different training methods and corpora (Fares et al. 2017). These embeddings were trained using the same corpora: English Wikipedia Dump (February 2017) and Gigaword $5^{th}$ Editio), containing about 6.8B tokens. The pre-processing steps on the corpora were identical with 10-word window size, lemmatization, and a vocabulary of 260K words.

Apart from the embeddings solely trained using contextual words, we also incorporated PARAGRAM (Wieting et al. 2015) in the experiment, which was initialized with Skip-gram embeddings and specialized with the paraphrases in PPDB (Ganitkevitch et al. 2013). In comparison with other popular embeddings (Yang and Yin 2021), PARAGRAM has achieved state-of-the-art results in lexical-semantic tasks. Note that PARAGRAM performed post-processing on SGNS using a triplet loss, which was equivalent to ATTRACT-REPEL except for exclusively using paraphrasable relationships in PPDB. PARAGRAM was trained using Wikipedia (about 1.8B tokens) with 300 dimensions.

### 4.2 Semantic constraints

To investigate the effects of semantic knowledge on enhancing distributional semantics, we selected two commonly used lexical resources: WordNet (Miller 1995, Fellbaum 1998) and Roget's thesaurus (Kipfer 2009). They contain about 909K and 555K synonymous pairs, respectively, along with 465K and 39K antonymous pairs. The dataset of direct hypernymy constraints on nouns and verbs consists of about 321K pairs, extracted only from WordNet's IS-A hierarchies, as there are no such relationships available in Roget's thesaurus.

### 4.3 Training setup

In line with ATTRACT-REPEL and PARAGRAM, we applied AdaGrad to optimize the cost functions in the above sections, through which specialization with each category of semantic constraint reached convergence after 20 epochs. For example, as with hierarchy-fitting, we ran a grid search to fine-tune the hyperparameter values, including learning rates, the margins of $m_{hyp}$, $m_{syn}$, and $m_{ant}$, and the size of mini-batch. Using the gold-standard data sets of WordSim-353-similarity (Agirre et al. 2009) on nouns (201 pairs) and the training part of SimVerb-3500 (Gerz et al. 2016) on verbs (500 pairs), we finalized the optimal hyper-parameter values: the learning rate of 0.03; the margins of $m_{hyp}$=0.6, $m_{syn}$=0.9, and $m_{ant}$=0.3; the regularizing factor of $\gamma_{reg} = 0.001$; and the mini-batch size of 128. In the quadruplet loss of hierarchy-fitting, the optimal values for $m_{hie\text{-}syn}$ and $m_{hie\text{-}hyp}$ were 0.001 and 0.6, respectively.

*4.4 Evaluation tasks*

We evaluated the specialized neural embeddings mainly on deriving semantic similarity from distributional semantics. Since we validated the specialized embeddings with semantic similarity datasets, and our semantic constraints included syn/antonymy and hypo/hypernymy, we employed two intrinsic tasks: judging semantic similarity and recognizing hypernymy to assess the post-processing methods.

4.4.1 Semantic similarity calculation

We validated hierarchy-fitting by reporting Spearman's rank correlation coefficient (ρ) on SimLex-999 (Hill et al. 2015), which contains nouns, verbs, and adjectives, as well as on the test part of SimVerb-3500 (Gerz et al. 2016) (denoted as SimVerb-3000), which is composed of 3,000 pairs of verbs. These gold standard datasets are often used to evaluate various methods in deriving semantic similarity (Yang and Yin 2021), mainly consisting of common words that frequently occur in generic domains. Moreover, we assessed hierarchy-fitting and other post-processing methods on the datasets collecting human similarity ratings on infrequent or rare words. We selected two benchmark datasets: RW-2034 (Luong et al. 2013) and CARD-660 (Pilehvar et al. 2018), for the evaluation.

Subject to the vocabulary size of the four embeddings, we tweaked a simple but effective back-off policy (Speer et al. 2017) to handle the issue of out-of-vocabulary (OOV) words. We deleted the last letter of an OOV word until its remaining string matched one of the tokens in the vocabulary. The simplified back-off policy can address the OOV issue in the evaluation with a full coverage rate on the datasets.

4.4.2 Hypernymy detection and directionality

Lexical entailment is a primary component of organizing concepts and inferring concept relationships in semantic memory (Collins and Quillian 1969). WordNet employed lexical entailment to interlink synsets or concepts into the IS-A hierarchies (Fellbaum and Miller 1990). Mining lexical entailment plays a critical role in knowledge engineering (Yang and Powers 2010, Navigli et al. 2011), e.g. automating taxonomy construction, enriching hand-crafted lexical resources, and augmenting downstream applications (Ido et al. 2013). Since hierarchy-fitting employs lexical entailment for semantic specialization, we evaluated it on detecting hypernymy and directionality, which proceeds in the same way as LEAR and LexSub.

Hypernymy identification in distributional semantics (Kiela et al. 2015, Nguyen et al. 2017) is often framed as an unsupervised learning task under two hypotheses: distributional inclusion (Weeds and Weir 2003, Geffet and Dagan 2005), claiming that a hypernym should share a substantial number of context features of its hyponyms; and distributional informativeness or generality (Santus et al. 2014), suggesting that a hypernym holds more generic or less informative features than its hyponyms. Unlike the symmetric synonymy and antonymy, hypernymy or its counterpart hyponymy is asymmetric and holds directionality in semantics. The hypotheses often work equivalently in identifying lexical entailment's directionality, essentially identical to information content (Resnik 1995). Information content supposed that a hypernym should be less informative or specific than its hyponyms for computing taxonomic similarity. In contrast to unsupervised learning on hypernymy identification, Levy et al. (2015) demonstrated that supervised learning might fail to recognize lexical entailment because it just tried to memorize distinctive contextual features of hyponyms or hypernyms rather than their underlying connections. Such lexical memorization could not deduce inherent relationships between a hyponym–hypernym pair because there was little connection information encoded in a distributional space. We, therefore, treated hypernymy detection and directionality as an unsupervised learning task, through which we can compare hierarchy-fitting with other post-processing methods.

LEAR defined a few metrics for encoding hypernymy directionality ($x_{hypo}$ to $x_{hyper}$), among which the following worked best:

$$L(x_{hypo}, x_{hyper}) = \frac{\|f_\theta(x_{hypo})\| - \|f_\theta(x_{hyper})\|}{\|f_\theta(x_{hypo})\| + \|f_\theta(x_{hyper})\|}$$

We, therefore, supplemented hierarchy-fitting NNEs with $L(x_{hypo}, x_{hyper})$ to distinguish the directionality of a hyponym–hypernym pair.

To thoroughly compare hierarchy-fitting with the specialization methods in Section 2, we used the HyperVec package (Nguyen et al. 2017) to evaluate them and report the average precision. The package defined an unsupervised metric of *HyperScore* for hypernymy identification, which was quantified as the product of a distributional similarity (*cosine*) between $x_{hypo}$ and $x_{hyper}$ and the ratio of Euclidean norms $x_{hypo}$ to $x_{hyper}$. *HyperScore* could indicate two critical aspects in recognizing lexical entailment: $x_{hypo}$ and $x_{hyper}$ should be distributionally similar at first; the Euclidean norm of $x_{hypo}$ should be less than it of $x_{hyper}$. The HyperVec package contains three benchmark datasets (Kiela et al. 2015) for detecting hypernymy and its directionality, which include:

1. **BLESS.** Word pairs in BLESS were annotated with various semantic relationships, from which 1,337 hyponym–hypernym pairs ($x_{hypo}$ and $x_{hyper}$) were extracted. The task for detecting hypernymy directionality in BLESS is to identify which word is a hypernym or superordinate simply by comparing the Euclidean norms of $x_{hypo}$ and $x_{hyper}$ in a ratio of $\|x_{hypo}\|$ to $\|x_{hyper}\|$. Supposing the $\|x_{hypo}\|$ should be shorter than $\|x_{hyper}\|$ on the premise that $x_{hyper}$ is more abstract or general than $x_{hypo}$, we can predict the directionality in a pair of $x_{hypo}$ and $x_{hyper}$ without setting up a predefined threshold.

2. **WBLESS.** WBLESS was designed for hypernymy detection, which contains 1,668 word pairs that were sourced from BLESS. A half of WBLESS was categorized as hypernymy or lexical entailment, and the other half with other semantic relationships such as hyponymy and meronymy. Following Nguyen et al. (2017), we formulated hypernymy detection in WBLESS as a binary classification task and fixed a threshold beforehand using 2% of WBLESS data, randomly extracted in 1,000 iterations.

3. **BIBLESS.** BIBLESS was a by-product of WBLESS, in which 834 hypernymy pairs were preserved, and the remaining 834 pairs were further grouped into two parts: 208 hyponymy pairs and 626 pairs of other relationships. BIBLESS is suitable for detecting both hypernymy and directionality, which is more complicated than detecting hypernymy on WBLESS and directionality on BLESS. We first attempted to locate a threshold to distinguish hypo/hypernymy from the other relationships and then searched for a second threshold to detect their directionality. We used the same procedure on WBLESS (Nguyen et al. 2017) to fix the thresholds.

4.4.3 Graded lexical entailment

Instead of considering hypernymy recognition as a binary classification task, Vulić et al. (2017) treated it as a regression task and created the HyperLex dataset for recognizing graded lexical entailment. They measured hypernymy or lexical entailment using a liker scale of similarity scores, assuming that lexical entailment should be appreciated gradually rather than binarily in human semantic memory (Collins and Quillian 1969). Apart from detecting hypernymy and directionality in the unsupervised binary classification tasks, we further validated hierarchy-fitting on HyperLex in deriving graded strength scores on lexical entailment.

There were 2,616 pairs of words in HyperLex, which were tagged with seven categories of semantic relations, such as hypernymy and meronymy. Each pair was rated with its association strength score, collected on a crowdsourcing platform. We first predicted *HyperScore* for each pair in HyperLex and then calculated Spearman's rank correlation coefficient to demonstrate how well each specialization method can approach humans on grading lexical entailment. The upper bound on HyperLex, a mean inter-annotator agreement, is 0.864. Note that among the seven semantic relations in HyperLex, hypernymy, containing different link distances in the IS-A hierarchies of WordNet, accounts for about 57.3%, whereas its counterpart, hyponymy, only holds about 11.3%.

## 5. Computing semantic similarity

Subject to its use of co-occurrence patterns, distributional semantics yield more semantic relatedness than semantic similarity. We aim to inject semantic relations to improve the efficacy of distributional semantics in deriving semantic similarity. We assess the specialization methods using the gold standard similarity datasets on common and rare words.

*5.1 Semantic similarity on common and rare words*

We first calculated distributional similarity between a pair of words and then measured Spearman's correlation between distributional similarity and average human ratings across each dataset. Table 2 shows the correlational results of using hierarchy-fitting and other post-processing methods in Section 2 to enhance neural embeddings. Given the Zipfian distribution (Zipf 1965) of word usage, we systematically compared these methods, using two classes of similarity datasets: SimLex-999 and SimVerb-3000 for frequent or common words and CARD-660 and RW-2034 for infrequent or rare words. As with SGNS, GloVe, and fastText, hierarchy-fitting almost achieved the best results on the four datasets. It improved the vanilla NNEs by 72.7% and 11.5% on average on the common- and rare-word datasets, respectively, whereas the state-of-the-art LEAR only attained 67.2% and 5.8%. Even on PARAGRAM, already specialized with the paraphrases in PPDB, hierarchy-fitting consistently outperformed other post-processing methods except on RW-2034.

Table 2: A comparison of measuring semantic similarity using different specialization measures on word embeddings. PG, SN, GV, and FT denote four embeddings: PARAGRAM, SGNS, GloVe, and fastText, respectively. Spearman's rank correlation coefficient ($\rho$) indicates the agreement between distributional similarity (*cosine*) and human similarity ratings. The score in the bracket stands for the upper bound of inter-annotator agreement (pairwise) on a dataset. The bold values indicate the best performance while imposing different semantic specializations.

|  | SimLex-999 (0.67) | | | | SimVerb-3000 (0.84) | | | |
| --- | --- | --- | --- | --- | --- | --- | --- | --- |
|  | PG | SN | GV | FT | PG | SN | GV | FT |
| Vanilla | 0.69 | 0.02 | 0.02 | 0.02 | 0.54 | 0.05 | 0.02 | 0.05 |
| Retrofitting | 0.68 | 0.19 | 0.21 | 0.19 | 0.55 | 0.16 | 0.16 | 0.16 |
| Counter-fitting | 0.74 | 0.28 | 0.27 | 0.28 | 0.63 | 0.22 | 0.19 | 0.21 |
| LEAR | 0.73 | 0.70 | 0.72 | 0.70 | 0.70 | 0.69 | 0.70 | 0.70 |
| Hierarchy-fitting | **0.82** | **0.75** | **0.76** | **0.78** | **0.77** | **0.75** | **0.75** | **0.75** |
|  | CARD-660 (0.89) | | | | RW-2034 (0.40) | | | |
|  | PG | SN | GV | FT | PG | SN | GV | FT |
| Vanilla | 0.24 | 0.19 | 0.19 | 0.17 | **0.41** | 0.28 | 0.23 | 0.30 |
| Retrofitting | 0.24 | 0.17 | 0.23 | 0.24 | 0.40 | 0.22 | 0.30 | 0.32 |
| Counter-fitting | 0.22 | 0.11 | 0.10 | 0.10 | **0.41** | 0.25 | 0.23 | 0.25 |
| LEAR | 0.20 | 0.23 | 0.24 | 0.25 | 0.26 | 0.33 | 0.33 | 0.33 |
| Hierarchy-fitting | **0.26** | **0.29** | **0.27** | **0.29** | 0.33 | **0.42** | **0.40** | **0.38** |

Note that LexSub only provided the specialized GloVe for evaluation, and its vanilla GloVe (with 300 dimensions) (Pennington et al. 2014) was trained with 6 billion tokens extracted from Wikipedia 2014 and Gigaword $5^{th}$ Edition corpora, with a vocabulary size of 400k. However, the vanilla GloVe in our experiment was trained with Wikipedia 2017 rather than 2014, with a vocabulary size of 260K. Given the potential impacts of using different corpora and vocabularies on yielding embeddings, we did not compare the LexSub results on GloVe with ours and only listed them for reference: 0.51 on SimLex-999; 0.40 on SimVerb-3000; 0.21 on CARD-660; 0.41 on RW-2034.

### 5.2 Salience of semantic relations

We also conducted an additive study on our hierarchy-fitting model to discriminate the validity of different semantic constraints, as shown in Table 3. Among mono-relations injected, all the semantic constraints substantially improved the vanilla embeddings on SimLex-999 and SimVerb-3000, but on the two rare-word datasets, they only showed a slight advance to the vanilla embeddings with 2.7% on average. As for the hybrid effect of adding semantic constraints incrementally on deriving semantic similarity, after injection of synonymy and antonymy, our model improved the vanilla embeddings by 57.4% on average on SimLex-999 and SimVerb-3000.

Table 3: A comparison of evaluating semantic constraints on specializing NNEs in the order of synonymy, antonymy, hypernymy, and hierarchy-fitting, along with additive evaluation on the relations.

|  | SimLex-999 | | | | SimVerb-3000 | | | |
| --- | --- | --- | --- | --- | --- | --- | --- | --- |
|  | PG | SN | GV | FT | PG | SN | GV | FT |
| Vanilla | 0.69 | 0.02 | 0.02 | 0.02 | 0.54 | 0.05 | 0.02 | 0.05 |
| Mono-synonymy | 0.77 | 0.68 | 0.67 | 0.68 | 0.71 | 0.65 | 0.64 | 0.64 |
| Mono-antonymy | 0.71 | 0.36 | 0.41 | 0.37 | 0.64 | 0.44 | 0.44 | 0.42 |
| Mono-hypernymy | 0.72 | 0.35 | 0.35 | 0.37 | 0.64 | 0.45 | 0.43 | 0.44 |
| Mono-hierarchy-fitting | 0.70 | 0.17 | 0.22 | 0.18 | 0.62 | 0.29 | 0.32 | 0.29 |
| + Injecting synonymy | 0.77 | 0.68 | 0.67 | 0.68 | 0.71 | 0.65 | 0.64 | 0.64 |
| + Injecting antonymy | **0.82** | 0.74 | 0.75 | 0.75 | **0.77** | 0.72 | 0.72 | 0.73 |
| + Injecting hypernymy | **0.82** | **0.79** | **0.79** | **0.79** | **0.77** | **0.75** | **0.75** | **0.76** |
| + Hierarchy-fitting | **0.82** | 0.75 | 0.76 | 0.78 | **0.77** | **0.75** | **0.75** | 0.75 |
|  | CARD-660 | | | | RW-2034 | | | |
|  | PG | SN | GV | FT | PG | SN | GV | FT |
| Vanilla | 0.24 | 0.19 | 0.19 | 0.17 | **0.41** | 0.28 | 0.23 | 0.30 |
| Mono-synonymy | 0.23 | 0.27 | 0.26 | 0.26 | 0.39 | **0.42** | **0.41** | **0.44** |
| Mono-antonymy | **0.26** | 0.19 | 0.21 | 0.16 | **0.41** | 0.31 | 0.31 | 0.35 |
| Mono-hypernymy | 0.25 | 0.25 | 0.23 | 0.24 | 0.33 | 0.23 | 0.20 | 0.26 |
| Mono-hierarchy-fitting | 0.25 | 0.22 | 0.21 | 0.21 | 0.37 | 0.25 | 0.24 | 0.27 |
| + Injecting synonymy | 0.23 | 0.27 | 0.26 | 0.26 | 0.39 | **0.42** | **0.41** | **0.44** |
| + Injecting antonymy | 0.21 | 0.25 | 0.23 | 0.24 | 0.33 | 0.40 | 0.39 | 0.41 |
| + Injecting hypernymy | **0.26** | 0.28 | 0.25 | 0.27 | 0.35 | 0.32 | 0.31 | 0.34 |
| + Hierarchy-fitting | **0.26** | **0.29** | **0.27** | **0.29** | 0.33 | **0.42** | 0.40 | 0.38 |

As for the two rare-word datasets, except for PARAGRAM, synonymy specialization augmented the three remaining vanilla embeddings by 11.7% on average, but continuously injecting antonymy was detrimental to the correlational results. Hypernymy specialization further enhanced NNEs in deriving semantic similarity, which reached their peak performances on SimLex-999 and SimVerb-3000 but not reliably on CARD-660 and RW-2034. When hierarchy-fitting the embeddings to further regularize hierarchical relationships, our model kept improving SGNS, GloVe, and fastText on CARD-660 and showed no significant change on RW-2034, although it deteriorated somewhat on SimLex-999 and remained nearly identical on SimVerb-3000.

*5.3 Summary*

The results of semantic specialization on NNEs emphasize the validity of hierarchy-fitting, which bears a close resemblance to ATTRACT-REPEL and LEAR in leveraging semantic relations to post-process distributional vectors. Hierarchy-fitting achieves state-of-the-art results in measuring semantic similarity. One of the key differences of our model from LEAR is that we exclusively employ direct rather than indirect hypernymy for semantic specialization. Semantic transitivity (Marathe and Hirst 2010) for indirect hypernymy may increase semantic distance, whereas semantic distance under direct hypernymy is relatively short and reliable to function as semantic constraints. Moreover, the hypernymy margin, a variable in the triplet loss, serves as the least similarity score for an authentic pair of direct hyponym–hypernym. If we keep it constant for any lexical entailment, mining indirect hypernymy to specialize NNEs may mistakenly attract and repel words in a distributional space. Hierarchy-fitting can enhance NNEs in deriving semantic similarity. It has further strengthened our hypothesis that collaborating multiple semantic relationships into one cost function may yield hierarchy embeddings.

## 6. Detecting lexical entailment

Specializing a distributional space with lexical-semantic relations can significantly enhance distributional semantics in deriving semantic similarity. We further evaluate different specialization approaches to recognizing lexical entailment in four tasks.

### 6.1 Unsupervised detection of hypernymy and directionality

We classified the specialization methods into two groups to investigate to what degree asymmetric distance (AD) metrics in late fusion could recognize hypernymy. In Table 4, Group 1 includes retrofitting, counter-fitting, LexSub, and hierarchy-fitting that employ no AD metrics. LexSub was placed in Group 1 as it only used affine transformation to distinguish hypo/hypernyms. Group 2 applies the AD metrics in Section 4.4.2 to hierarchy-fitting mining two variants of hypernyms: one only contains direct hypernymy with the size of 0.32M, denoted as AD_*dir*; and the other consists of multiple-level or indirect hypernymy with the size of 1.55M, denoted as AD_*indir*. Note that the size of AD_*indir* is the same as it in LEAR. Apart from the post-processing methods in Section 2, Table 4 includes HyperVec (Nguyen et al. 2017), a state-of-the-art joint training method in early fusion for detecting hypernymy. We evaluated hierarchy-fitting in Group 2 with AD_*dir* and AD_*indir*, as shown in Tabel 4.

In Table 4, no specialization methods in Group 1 show particular advantages over others, with retrofitting scoring about the best precision of 0.57 (fastText) on BLESS, and counter-fitting about 0.54 and 0.38 (PARAGRAM) on WBLESS and BIBLESS, respectively. Hierarchy-fitting performed competitively across the three datasets. After specialization on NNEs with the asymmetric distance metric, Hierarchy-fitting+AD_*dir* improved hierarchy-fitting by 0.35, 0.13, and 0.08 on BLESS, WBLESS, and BIBLESS, respectively; and Hierarchy-fitting+AD_*indir* further boosted Hierarchy-fitting+AD_*dir* by 0.32, 0.26, and 0.35. Hierarchy-fitting+AD_*indir* was barely distinguishable from state-of-the-art methods, including LEAR and HyperVec (Nguyen et al. 2017). It significantly surpassed the initial results of Kiela et al. (2015) on the three datasets, although Kiela et al. (2015) employed image generality rather than linguistic generality of hypernyms in the evaluation. Note that in learning a separate space for hypernym detection, LexSub defined an asymmetric distance function through the affine transformation of embeddings. However, LexSub only achieved 0.34 (BLESS), 0.47 (WBLESS), and 0.33 (BIBLESS) in our experiments, which were deviated from its initial results of 0.21, 0.60, and 0.50 achieved using the *cosine* similarity only.

Table 4: Results of detecting hypernymy and directionality on three benchmark datasets.

|  | BLESS | | | | WBLESS | | | | BIBLESS | | | |
|---|---|---|---|---|---|---|---|---|---|---|---|---|
|  | PG | SN | GV | FT | PG | SN | GV | FT | PG | SN | GV | FT |
| Vanilla | 0.50 | 0.43 | 0.44 | 0.32 | 0.50 | 0.48 | 0.51 | 0.46 | 0.35 | 0.34 | 0.37 | 0.33 |
| Retrofitting | 0.37 | 0.53 | 0.45 | 0.57 | 0.48 | 0.51 | 0.47 | 0.52 | 0.34 | 0.36 | 0.34 | 0.37 |
| Counter-fitting | 0.54 | 0.34 | 0.43 | 0.50 | 0.54 | 0.48 | 0.48 | 0.50 | 0.38 | 0.35 | 0.35 | 0.36 |
| LexSub |  |  | 0.34 |  |  |  | 0.47 |  |  |  | 0.33 |  |
| Hierarchy-fitting | 0.28 | 0.29 | 0.29 | 0.30 | 0.48 | 0.47 | 0.48 | 0.47 | 0.36 | 0.35 | 0.36 | 0.35 |
| Hierarchy-fitting+AD_*dir* | 0.71 | 0.64 | 0.60 | 0.62 | 0.65 | 0.62 | 0.58 | 0.58 | 0.47 | 0.47 | 0.43 | 0.38 |
| Hierarchy-fitting+AD_*indir* | 0.96 | 0.96 | 0.97 | 0.97 | 0.86 | 0.87 | 0.86 | 0.86 | 0.79 | 0.80 | 0.79 | 0.78 |
| LEAR | 0.96 | 0.96 | 0.96 | 0.96 | 0.89 | 0.88 | 0.88 | 0.89 | 0.85 | 0.84 | 0.84 | 0.85 |
| HyperVec (Nguyen et al. 2017) | 0.92 |  |  |  | 0.87 |  |  |  | 0.81 |  |  |  |
| Kiela et al. (2015) | 0.88 |  |  |  | 0.75 |  |  |  | 0.57 |  |  |  |

Our results reflect that leveraging asymmetric distance to encode the directionality of lexical entailment can enhance distributional semantics in hypernymy identification, which corroborates well with LEAR (Vulic and Mrkšić 2018). It also confirms the previous findings of Levy et al. (2015) on the limitations of unaltered distributional semantics. Our study provides further evidence that calculating distributional similarity with *cosine* is appropriate for detecting symmetric relations or deriving semantic similarity but not for detecting asymmetric relations such as IS-A and PART-OF relations. In the same vein of LEAR, the directionality of hypernymy can also be recognized on the hypotheses of distributional inclusion (Weeds and Weir 2003, Geffet and Dagan 2005) and generality (Santus et al. 2014). For example, distributional inclusion leveraged the intersection of contextual features sharing between hypo/hypernyms for identifying lexical entailment (Weeds and Weir 2003, Weeds et al. 2004, Kotlerman et al. 2010) or for joint training in late fusion to yield hierarchical embeddings (Nguyen et al. 2017); and distributional generality used information entropy to detect the less informative contextual features of hypernyms. Without imposing directionality specialization on distributional semantics, there may be little tangible information that can be captured to represent lexical entailment.

Further analysis indicated that the remarkable gain of Hierarchy-fitting+AD_*indir* and LEAR on recognizing hypernymy might be at the expense of their deteriorating outcomes on measuring semantic similarity, as shown in Figure 2. Hierarchy-fitting+AD_*dir* only used direct IS-A relations to enhance distributional semantics on measuring semantic similarity and distinguishing lexical entailment, and the results indicated that it matched well with hierarchy-fitting in balancing different semantic constraints. Injecting more indirect IS-A relations was detrimental to hierarchy-fitting+AD_indir and LEAR in deriving semantic similarity, adversely affecting semantic specialization on distributional semantics. This result has further supported our hypothesis that semantic transitivity caused by introducing multiple-level of lexical entailment may impose improper constraints on regularizing a distributional space.

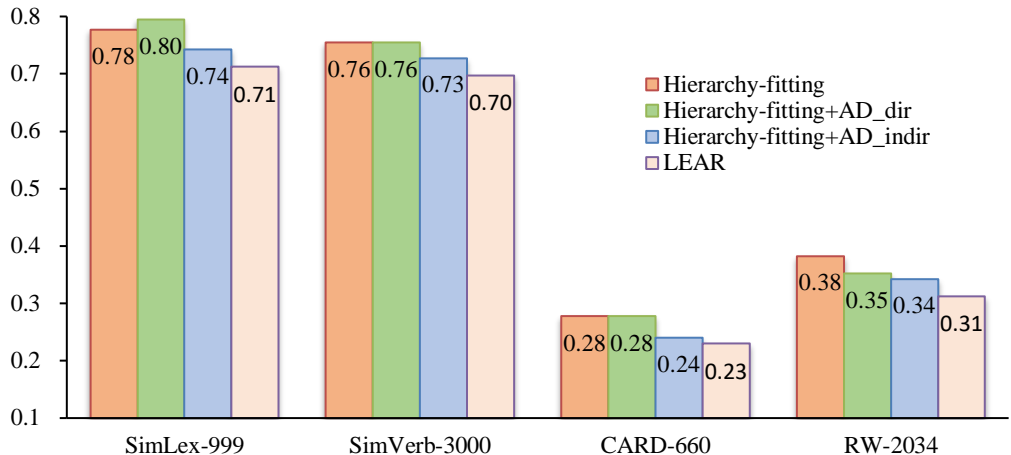

Figure 2: Spearman correlation results of Hierarchy-fitting with/out the asymmetric distance function on measuring semantic similarity. A mean correlation on the four benchmark datasets on the NNEs: PG, SN, GV, and FT was calculated for each method.

## 6.2 Graded lexical entailment

We assessed hierarchy-fitting in identifying hypernymy and directionality as a binary classification task. On the other hand, as Vulić et al. (2017) has highlighted in HyperLex, apprehension of lexical entailment might exist as a gradual process in human semantic memory, and lexical entailment should be quantified with a graded or continuous value. We, therefore, further compare the semantic specialization methods in measuring graded lexical entailment in HyperLex.

As shown in Table 5, without using any asymmetric distance to specify hypernymy's directionality, hierarchy-fitting attained best on PG, SN, GV, and FT in Group 1, whereas other methods, including retrofitting, counter-fitting, and LexSub, almost showed no improvement on each vanilla embedding. After factoring in the asymmetric character of hypernymy, hierarchy-fitting+AD_*dir* attained moderate improvements compared to hierarchy-fitting; and with the injection of more indirect IS-A relations, hierarchy-fitting+AD_*indir* further advanced hierarchy-fitting+AD_*dir* by 0.30 on average.

Among the state-of-the-art methods in refining distributional spaces with asymmetric hypernymy, LEAR best measured graded lexical entailment but fell behind hierarchy-fitting in deriving semantic similarity, as shown in Figure 2. HyperVec employed both direct and indirect IS-A relations in joint-training NNEs and was inferior to LEAR and hierarchy-fitting+AD_*indir*. LexSub created a separate subspace for hypernymy detection but gained a worsening result in HyperLex. These results indicated that adding asymmetric distance to generate the directionality of lexical entailment in late fusion might be more effective than both joint training in early fusion (Nguyen et al. 2017) and singling out a distributional sub-space dedicated to hypernymy (Arora et al. 2020).

Table 5 : Spearman correlation results of different specialization methods in computing graded lexical entailment on HyperLex.

|  | PG | SN | GV | FT |
|---|---|---|---|---|
| Vanilla | 0.24 | 0.23 | 0.21 | 0.13 |
| Retrofitting | 0.22 | 0.10 | 0.10 | 0.09 |
| Counter-fitting | 0.32 | 0.21 | 0.16 | 0.21 |
| LexSub |  |  | 0.18 |  |
| Hierarchy-fitting | 0.35 | 0.36 | 0.36 | 0.36 |
| Hierarchy-fitting+AD_*dir* | 0.38 | 0.39 | 0.38 | 0.38 |
| Hierarchy-fitting+AD_*indir* | 0.69 | 0.68 | 0.69 | 0.66 |
| LEAR | 0.71 | 0.71 | 0.71 | 0.70 |
| HyperVec: (Nguyen et al. 2017) | 0.54 |  |  |  |

### 6.3 Summary

Our semantic specialization method, hierarchy-fitting, can achieve state-of-the-art results in recognizing hypernymy in both binary and gradual evaluations. Since the evaluation datasets contain direct and indirect IS-A relations, we inject them to enhance NNEs through learning an asymmetric distance function to encode hypernymy's directionality. Overall, the advantage of hierarchy-fitting over other specialization methods such as LEAR and LexSub is that it can leverage multiple semantic constraints in detecting hypernymy and directionality without compromising semantic similarity computation.

## 7. Conclusion

Semantically specializing neural word embeddings aims to aggregate lexical and distributional semantics in early or late fusion. We have proposed to train a distance metric in late fusion to refine a distributional space to pursue the goal. Together, in a specialized space, metric learning can attract synonyms closer, repel antonyms farther, and pull hypo/hypernyms closer while maintaining hypernymy's directionality. Moreover, distributional semantics can be maximumly preserved when merging semantic constraints' contributions in loss functions. Our hierarchy-fitting method defined a quadruplet loss to regularize semantic difference among symmetric syn/antonymy and asymmetric IS-A relations in specializing neural embeddings, which can significantly improve distributional semantics in deriving semantic similarity and recognizing lexical entailment. This investigation corroborates those of earlier studies on semantic specialization on distributional vectors. It is scalable to other tasks, e.g. training a vector space model for measuring semantic relatedness and detecting mero/holonymy.

Post-processing methods can significantly reduce the computational workload caused by unsupervised joint-training processes to yield semantics-enhanced neural embeddings. To further improve such late fusion methods and enrich the semantic content of distributional vectors, we recommend that the follow-up phase of the study should concentrate on the following:

1. Training a unified distance metric while injecting different linguistic constraints. An integrated loss function can simplify hyperparameters' tuning process in metric learning and improve the generalization of semantics-enhanced distributional vectors. The margins in the metric should be adjustable according to types of semantic relations. Hierarchy-fitting can apply a different margin to indirect hypernymy, distinguishing it from direct hypernymy to inherently reflect semantic nuances when taxonomic links of lexical entailment vary. A fixed margin will incur pushing in/direct hypernymy within the same distance range, inevitably miscalculating their semantic distance. A more flexible way of setting up margins is to dynamically adjust their values according to taxonomic distances on different semantic relations.

2. Refining a distributional space through multimodal semantic constraints. The present studies mainly focus on examining the linguistic aspect of semantic constraints, such as lexical semantics from WordNet, Roget's Thesaurus, and PPDB. Cross-lingual concept relations from BabelNet is also helpful to yield a unified embedding for different languages, given the unbalanced distribution of linguistic resources. Besides linguistic constraints, semantic content can be grounded on other perceptual sources such as vision and sound, e.g. hypernym detection using image generality (Kiela et al. 2015). Such heterogeneous data are complementary to constructing a robust and unified data representation in a multimodal semantic space (Bruni et al. 2014). Deep metric learning via late fusion can enrich a distributional space while lowering the complexity of early information fusion on symbolic and signal information. Doing so can harvest the synergy from mining hand-crafted knowledge resources, multimodal data, and the combination thereof.

## Acknowledgements

The work has been supported by National Social Science Foundation (China) (Grant No. 17BYY119).